\documentclass[10pt,twocolumn,letterpaper]{article}

\usepackage{cvpr}

\usepackage{times}
\usepackage{epsfig}
\usepackage{graphicx}
\usepackage{amsmath}
\usepackage{amssymb}

\usepackage{booktabs}
\usepackage{multirow}
\usepackage{float}
\usepackage[autolanguage]{numprint}
\usepackage{lipsum}

\usepackage[pagebackref=true,breaklinks=true,letterpaper=true,colorlinks,bookmarks=false]{hyperref}

\cvprfinalcopy 

\def\cvprPaperID{} 

\begin{document}

\title{Visual Entailment: A Novel Task for Fine-Grained Image Understanding}

\author{Ning Xie\thanks{Work performed as a NEC Labs intern}\\
Wright State University\\
Dayton, OH, U.S.A.\\
{\tt\small xie.25@wright.edu}
\and
Farley Lai\\
NEC Laboratories America\\
Princeton, NJ, U.S.A.\\
{\tt\small farleylai@nec-labs.com}
\and
Derek Doran\\
Wright State University\\
Dayton, OH, U.S.A.\\
{\tt\small derek.doran@wright.edu}
\and
Asim Kadav\\
NEC Laboratories America\\
Princeton, NJ, U.S.A.\\
{\tt\small asim@nec-labs.com}
}

\maketitle

\begin{abstract}

Existing visual reasoning datasets such as Visual Question Answering (VQA),
often suffer from biases conditioned on the question, image or answer
distributions. The recently proposed CLEVR dataset addresses 
these limitations and requires fine-grained reasoning but the dataset is synthetic and consists of similar objects and sentence structures across the dataset.

In this paper, we introduce a new inference task,
{\bf Visual Entailment} (VE) - consisting of image-sentence pairs
whereby a premise is defined by an image, rather than a natural language sentence as in traditional Textual Entailment tasks. The goal of a trained VE model is to predict whether the image semantically entails the text.
To realize this task, we build a dataset SNLI-VE based on the Stanford Natural Language Inference corpus and Flickr30k dataset. We evaluate various existing VQA baselines and build a model called Explainable Visual Entailment (EVE) system to address the VE task. EVE achieves up to 71\% accuracy and outperforms several other state-of-the-art VQA based models. Finally, we demonstrate the explainability of EVE through cross-modal attention visualizations.
The SNLI-VE dataset is publicly available at \url{https://github.com/necla-ml/SNLI-VE}.

\end{abstract}
\section{Introduction}


The pursuit of ``visual intelligence'' is a long lasting theme of the machine learning community.  While the performance of image classification and object detection has significantly improved
in the recent years~\cite{krizhevsky2012imagenet, simonyan2014very, szegedy2015going, he2016deep},
progress in higher-level scene reasoning tasks such as scene understanding is relatively limited~\cite{wu2017visual}.

Recently, several datasets, such as VQA-v1.0~\cite{antol2015vqa}, VQA-v2.0~\cite{goyal2017making}, CLEVR~\cite{johnson2017clevr}, Visual7w~\cite{zhu2016visual7w}, Visual Genome~\cite{krishnavisualgenome}, COCO-QA~\cite{ren2015exploring}, 
and models~\cite{johnson2017inferring, santoro2017simple, hudson2018compositional, jiang2018pythia, anderson2018bottom, teney2017tips, fukui2016multimodal, kim2018bilinear} have been used to measure the progress in understanding the interaction between vision and language modalities.
However, the quality of the widely used VQA-v1.0 dataset~\cite{antol2015vqa}
suffers from a natural bias~\cite{goyal2017making}. 
Specifically, there is a long tail distribution of answers and also a question-conditioned bias where, questions may hint at the answers, such that the correct answer may be inferred without even considering the visual information. 
For instance, of the question ``Do you see a \ldots ?'', the model may bias towards the answer ``Yes'' since it is correct for 87\% of times during training.
Besides, many questions in the VQA-v1.0 dataset are simple and straightforward and do not require compositional reasoning from the trained model.
VQA-v2.0~\cite{goyal2017making} has been proposed to reduce the dataset ``bias'' considerably in VQA-v1.0 by associating each question with relatively balanced different answers. However, the questions are rather straight-forward and require limited fine-grained reasoning.

CLEVR dataset~\cite{johnson2017clevr}, is designed for fine-grained reasoning and consists of compositional questions such as ``What size is the cylinder that is left of the brown metal thing that is left of the big sphere?''. This kind of questions requires learning fine-grained reasoning based on visual information. However, CLEVR is a synthetic dataset, and visual information and sentence structures are very similar across the dataset. 
Hence, models that provide good performance on CLEVR dataset may not generalize to real-world settings.

\begin{figure}[h]
  \centering
  \includegraphics[width=\linewidth]{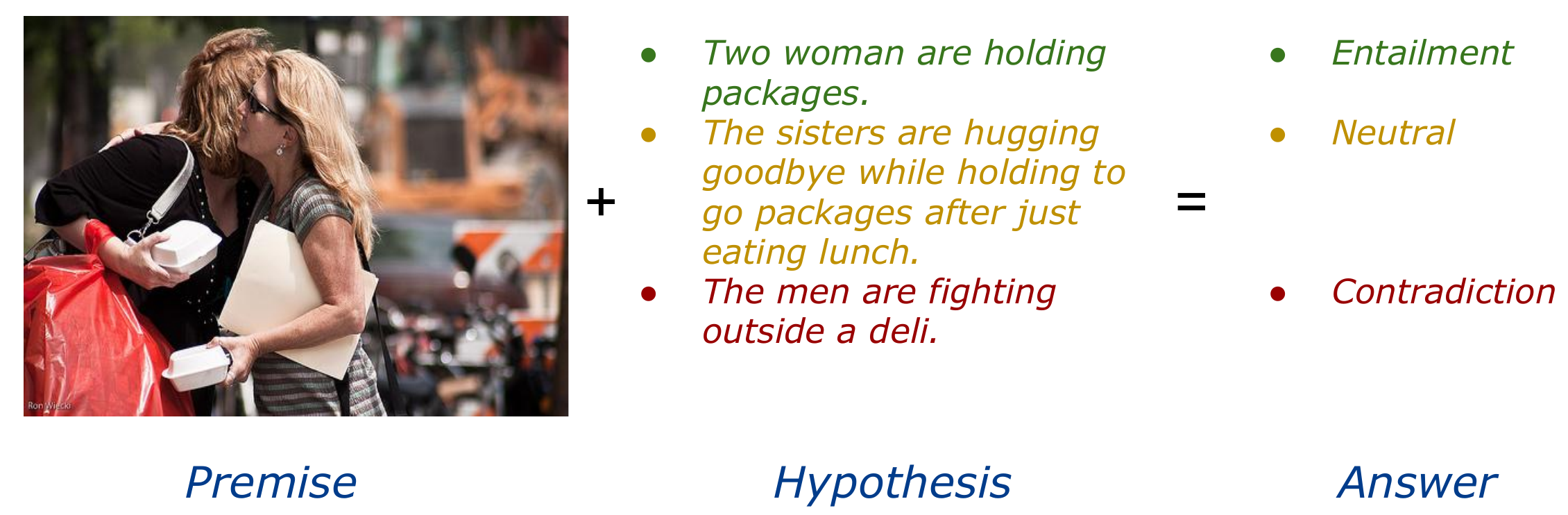}
  \caption{An Example from SNLI-VE dataset}
  \label{fig:SNLI-VE-eg1}
\end{figure}

To address the above limitations, we propose a novel inference task, {\em Visual Entailment} (VE), which requires fine-grained reasoning in real-world settings. 
The design is derived from Text Entailment (TE)~\cite{dagan2006pascal} task.
In our VE task, a real world image premise $P_{image}$ and a natural language hypothesis $H_{text}$ are given, and the goal is to determine if $H_{text}$ can be concluded given the information provided by $P_{image}$. Three labels {\em entailment}, {\em neutral} or {\em contradiction} are assigned based on the relationship conveyed by the $(P_{image}, H_{text})$.
\begin{itemize}
    \item \emph{Entailment} holds if there is enough evidence in $P_{image}$ to conclude that $H_{text}$ is true.
    \item \emph{Contradiction} holds if there is enough evidence in $P_{image}$ to conclude that $H_{text}$ is false.
    \item Otherwise, the relationship is \emph{neutral}, implying the evidence in $P_{image}$ is insufficient to draw a conclusion about $H_{text}$.
\end{itemize}
The main difference between VE and TE task is, the premise in TE in a natural language sentence $P_{text}$, instead of an image premise $P_{image}$. Note that the existing of ``neutral'' makes the VE task more challenging compared to previous ``yes-no'' VQA tasks, since ``neutral'' requires the model to conclude the uncertainty between ``entailment (yes)'' and ``contradiction (no)''. 
Figure~\ref{fig:SNLI-VE-eg1} illustrates a VE example, which is from the SNLI-VE dataset we propose below, that given an image premise, the three different text hypotheses lead to different labels.

We build the SNLI-VE dataset to illustrate the VE task, based on Stanford Natural Language Inference (SNLI)~\cite{bowman2015large}, which is a widely used text-entailment dataset, and Flickr30k~\cite{young2014image}, which is an image captioning dataset. 
The combination of SNLI and Flickr30k is straightforward since SNLI is created using Flickr30k. The detailed process of creating the SNLI-VE dataset is discussed in Section~\ref{subsec:SNLI-VE-dataset}.

We develop an Explainable Visual Entailment (EVE) model to address the VE task. 
EVE captures the interaction within and between the image premise and the text hypothesis through attention.
We evaluate EVE against several other state-of-the-art (SOTA) visual question answering (VQA) baselines and an image captioning based model on the SNLI-VE dataset.
The interpretability of EVE is demonstrated using attention visualizations.

In summary, the contributions of our work are:
\begin{itemize}
    \item We propose a {\em novel} inference task, Visual Entailment, that requires a systematic cross-modal understanding between vision and a natural language.
    \item We build a VE dataset, SNLI-VE, consisting of real-world image and natural language sentence pairs 
    for VE tasks.
    The dataset is publicly available\footnote{\url{https://github.com/necla-ml/SNLI-VE}}.
    \item We design a VE model, EVE, to solve the VE task with interpretable attention visualizations.
    \item We evaluate EVE against other SOTA VQA and image captioning based baselines.
\end{itemize}

\section{Related Work}
\label{sec:related}

Our work is inspired by previous work on NLI, VQA, image captioning, and interpretable models.


\paragraph{Natural Language Inference.} 
We focus on \emph{textual entailment} as our NLI task~\cite{fyodorov2000natural, condoravdi2003entailment, bos2005recognising, dagan2006pascal, maccartney2009extended}. Annotated corpus for TE was limited in size until SNLI \cite{bowman2015large} was proposed, which is based on the Flickr30k \cite{young2014image} image captions. Since then, several neural-network based methods have been proposed over SNLI that either use sentence encoding models to individually encode hypothesis and premise or attention based models that encode the sentences together and align similar words in hypothesis and premise~\cite{chen2016enhanced, nie2017shortcut, shen2018reinforced, rocktaschel2015reasoning}.
Our paper extends the TE task in the visual domain -- allowing future work on our SNLI-VE task to build new models on recent progress in SNLI and VQA. 
Our work is different from the recent work \cite{vu2018grounded} that combines both images and captions as premises.

\paragraph{Visual Question Answering.}
Recent work on VQA includes datasets \cite{johnson2017clevr, antol2015vqa, goyal2017making, zhu2016visual7w, krishnavisualgenome, ren2015exploring, malinowski2014multi, gao2015you, tapaswi2016movieqa} and models \cite{johnson2017inferring, santoro2017simple, hudson2018compositional, jiang2018pythia, anderson2018bottom, teney2017tips, fukui2016multimodal, kim2018bilinear}. 
The goal of VQA is to answer natural language questions based on the provided visual information.
VQA-v2.0 \cite{goyal2017making} and CLEVR \cite{johnson2017clevr} datasets are designed to address bias and reasoning limitations of VQA-v1.0, respectively.
Recent work on compositional reasoning systems have achieved nearly 100\% results on CLEVR ~\cite{hudson2018compositional} 
but the SOTA performance on VQA-v2.0 is no more than 75\% \cite{vqa2018leaderboard},
implying learning multi-modal feature interaction using natural images has room for improvement.
There have been a large number of models and approaches to address the VQA task. 
This includes simple linear models using ranking loss \cite{frome2013devise, karpathy2015deep},
bi-linear pooling methods~\cite{lin2015bilinear, gao2016compact, pham2013fast, fukui2016multimodal, kim2018bilinear},
attention-based methods ~\cite{anderson2018bottom, pedersoli2016areas, singh2018attention}
and reasoning based approaches~\cite{perez2017film, hu2017learning, johnson2017inferring, kim2018progressive, hudson2018compositional} on CLEVER and VQA-v1.0 datasets.

\paragraph{Image Captioning.} 
The problem of image captioning explores the generation of natural language sentences to best depict input image content. 
A common approach for these tasks is to use temporal models over convolutional features~\cite{karpathy2015deep, vinyals2017show, chen2017sca}. 
Recent work has also explored generating richer captions to describe images in a more fine-grained manner~\cite{johnson2016densecap}. 
EVE differs from image-captioning since it requires discerning fine-grained information about an image conditioned on the hypothesis into three classes. 
However, existing image-captioning methods can serve as a baseline, where the output class label is based on a distance measure between the generated caption and the input hypothesis.

\paragraph{Visual Relationship Detection.}
Relationship detection among image constituents uses separate branches in a ConvNet to model objects, humans, and their interactions ~\cite{chao2017learning, gkioxari2017detecting}.
A distinct approach in Santoro et al. ~\cite{santoro2017simple} treats each of the cells across channels in convolutional feature maps as an object and the relationships are modeled by a pairwise concatenation of the feature representations of individual cells. 

Scene graph based relationship modeling, using a structured representation for describing object relationships and their attributes ~\cite{johnson2015image, li2017scene, liang2017deep, xu2017scene} has been extensively studied. 
Furthermore, pairing different objects in a scene~\cite{dai2017detecting, hu2016modeling, santoro2017simple, zhang2017visual} is also common.
However, a scene with many objects may have only a few individual interacting objects. 
Hence, it can be inefficient to model all relationships across all individual object pairs~\cite{zhang2017relationship}, making these methods computationally expensive for complex scene understanding tasks such as VE.

Our model, EVE instead uses self-attention to efficiently learn the relationships between various scene elements and words instead of bi-gram or tri-gram based modeling as used in previous work.

\paragraph{Interpretability.} 
As deep neural networks have become widespread in real-world applications, there has been an increasing focus on interpretability and transparency. 
Recent work addresses this requirement either through saliency-map visualizations ~\cite{selvaraju2017grad, zeiler2014visualizing, montavon2017explaining}, attention mechanism ~\cite{xu2015show, zhang2018top,  park2016attentive, das2017human}, or other analysis ~\cite{jiang2018trust, koh2017understanding, raghu2017svcca, ribeiro2016should}. 
Our work demonstrates interpretability via attention visualizations.

\section{Visual Entailment Task}
\label{sec:vet}

\begin{figure}[h]

  \begin{minipage}[t]{\columnwidth}
  \centering
  \includegraphics[width=\linewidth]{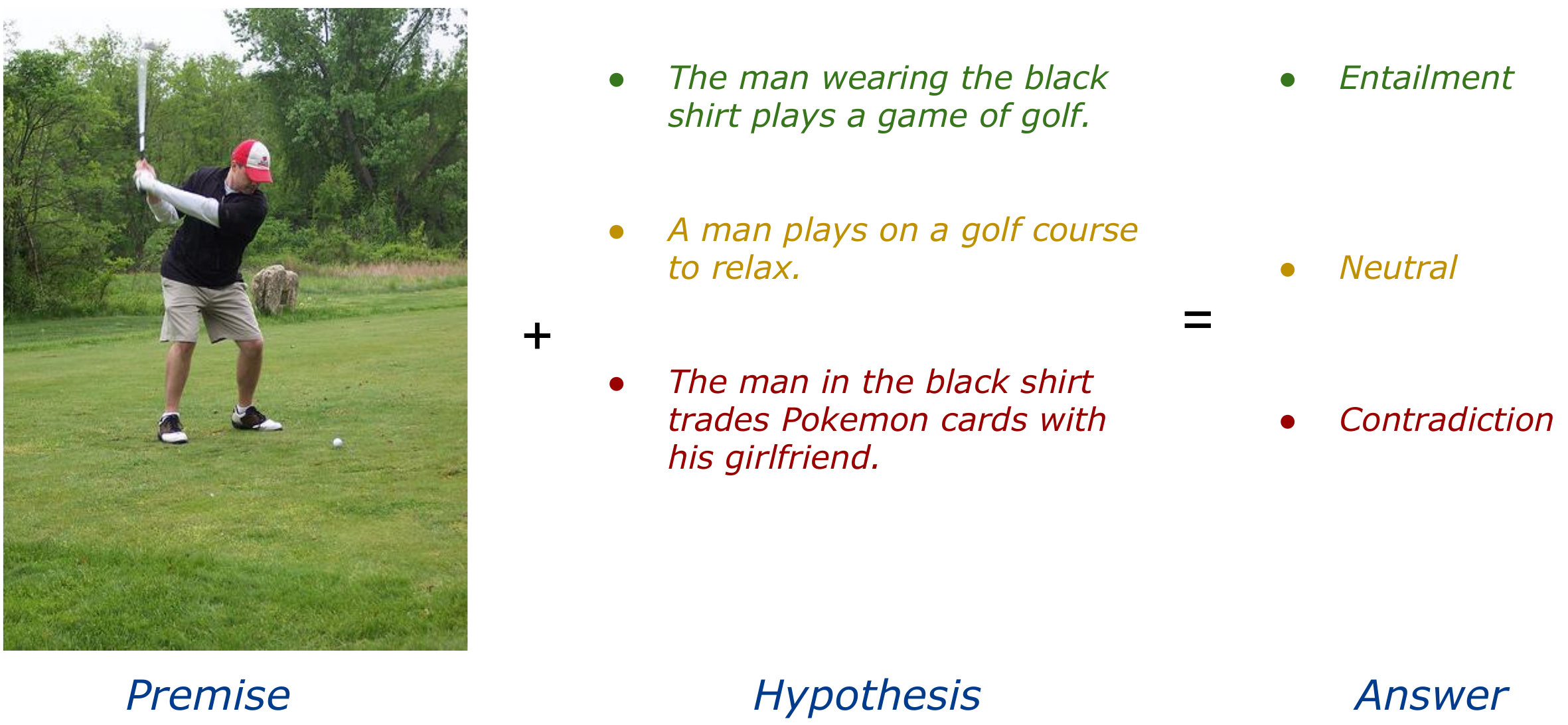}
  \end{minipage}

  \begin{minipage}[t]{\columnwidth}
  \centering
  \includegraphics[width=\linewidth]{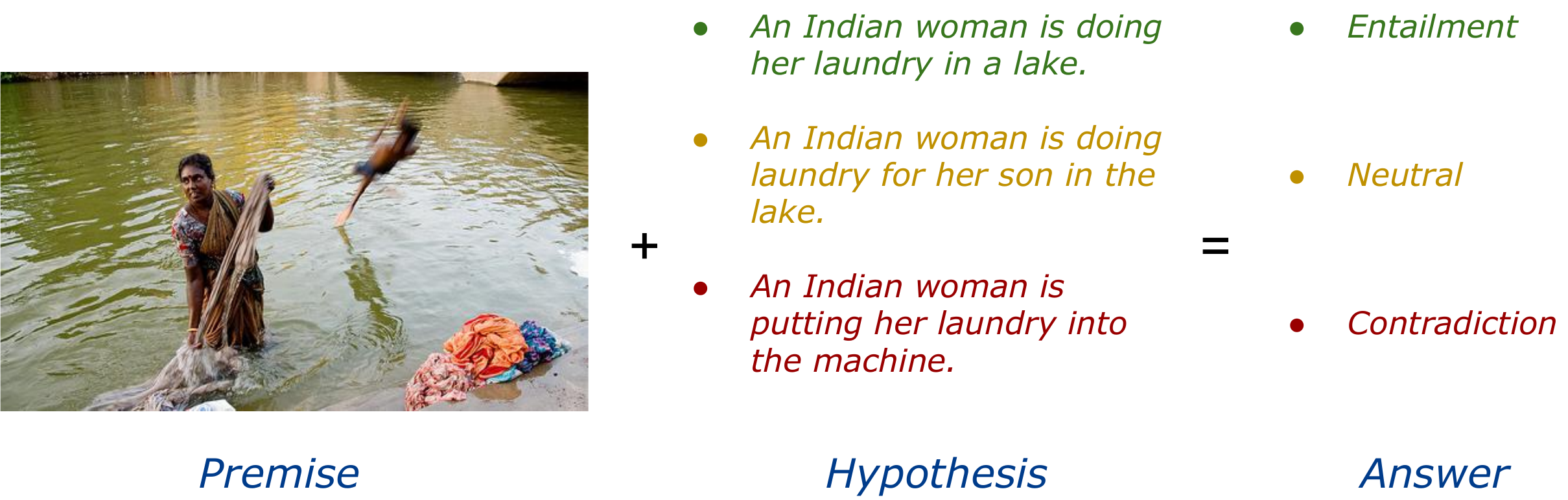}
  \end{minipage}
  
  \begin{minipage}[t]{\columnwidth}
  \centering
  \includegraphics[width=\linewidth]{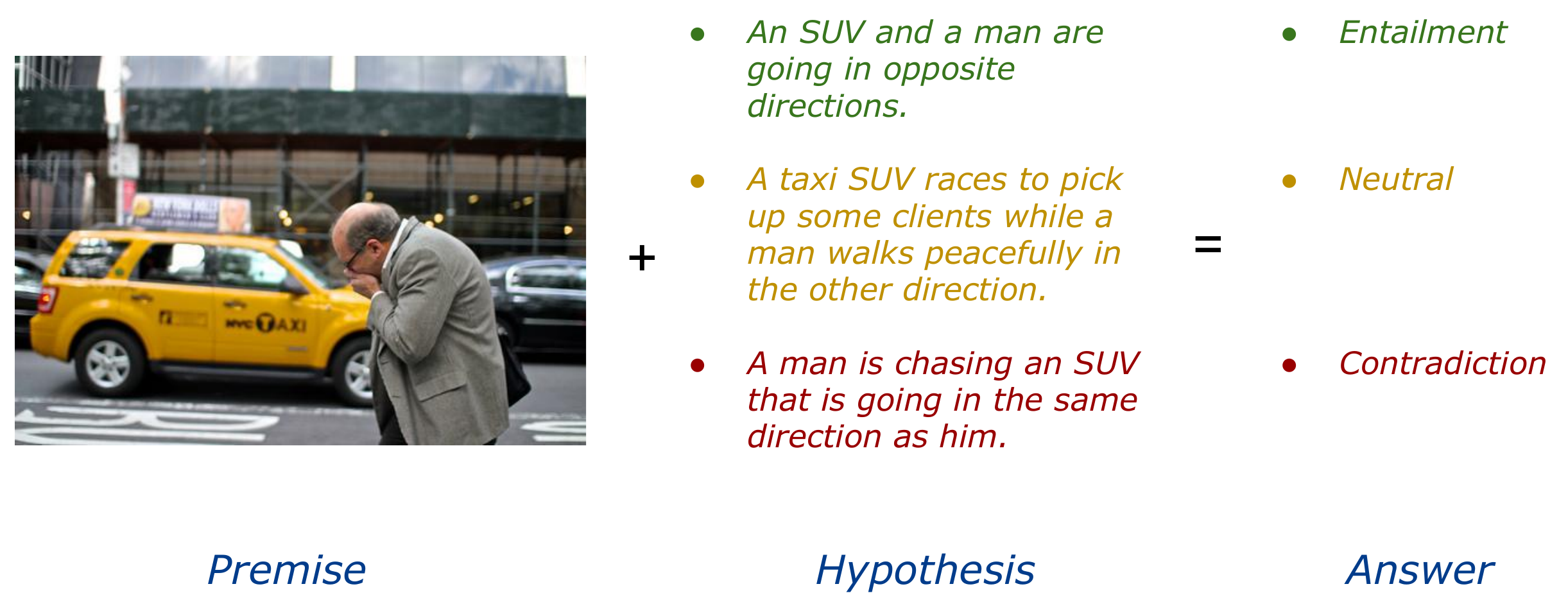}
  \end{minipage}

\caption{\bf More examples from SNLI-VE dataset}
\label{fig:SNLI-VE-more-egs}
\end{figure}

\subsection{Formal Definition} \label{subsec:formal-definition}

We introduce a dataset $\mathcal{D}$  for VE task structured as 
${\{(i_1, h_1, l_1), (i_1, h_2, l_2) \dots (i_1, h_{m_1}, l_{m_1}), \dots (i_n, h_{m_n}, l_{m_n})\}}$, where $(i_k, h_s, l_s)$ is an instance from $\mathcal{D}$, with $i_k$, $h_s$, and $l_s$ denoting an image premise, a text hypothesis and a class label, respectively. It is worth noting that each image $i_k$ is used multiple times with different labels given distinct hypotheses $\{h_{m_k}\}$.

Three labels $e$, $n$, or $c$ are assigned based on the relationship conveyed by $(i_k,h_s)$. 
Specifically, 
i) $e$ (entailment) is assigned if $i_k \models h_s$, 
ii) $n$ (neutral) is assigned if $i_k \not\models h_s \land i_k \not\models \lnot h_s$, 
iii) $c$ (contradiction) is assigned if $i_k \models \lnot h_s$.

\subsection{Visual Entailment Dataset}
\label{subsec:SNLI-VE-dataset}

\subsubsection{Dataset criteria}
Based on the vision community's experience with SNLI, VQA-v1.0, VQA-v2.0, and CLEVR, there are four \emph{criteria} in developing an effective dataset:
\begin{enumerate}

\item \emph{Structured set of real-world images}. The dataset should be based on real-world images and the same image can be paired with different hypotheses to form different labels.

\item \emph{Fine-grained}. The dataset should enforce fine-grained reasoning about subtle changes in hypotheses that could lead to distinct labels.

\item \emph{Sanitization}. No instance overlapping across different dataset partitions. One image can only exist in a single partition. 

\item \emph{Account for any bias}. Measure the dataset bias and provide baselines to serve as the performance lower bound for potential future evaluations.

\end{enumerate}

\subsubsection{SNLI-VE Construction}
We now describe how we construct SNLI-VE, which is a dataset for VE tasks.

We build the dataset SNLI-VE based on two existing datasets, Flickr30k ~\cite{young2014image} and SNLI ~\cite{bowman2015large}.
\textbf{Flickr30k} is a widely used image captioning dataset containing 31,783 images and 158,915 corresponding captions.
The images in Flickr30k consist of everyday activities, events and scenes ~\cite{young2014image}, with 5 captions per image generated via crowdsourcing.
\textbf{SNLI} is a large annotated TE dataset built upon Flickr30k captions.
Each image caption in Flickr30k is used as a text premise in SNLI.
The authors of SNLI collect multiple hypotheses in the three classes - \emph{entailment}, \emph{neutral}, and \emph{contradiction} - for a given premise via Amazon Mechanical Turk \cite{turk2012amazon}, resulting in about 570K $(P_{text}, H_{text})$ pairs. Data validation is conducted in SNLI to measure the label agreement. 
Specifically, each $(P_{text}, H_{text})$ pair is assigned a \emph{gold label}, indicating the label is agreed by a majority of crowdsourcing workers (at least 3 out of 5).
If such a consensus is not reached, the gold label is marked as ``-''.

Since SNLI was constructed using Flickr30k captions, for each $(P_{text}, H_{text})$ pair in SNLI, it is feasible to find the corresponding Flickr30k image through the annotations in SNLI. 
This enables us to create a structured VE dataset based on both.
Specifically, for each $(P_{text}, H_{text})$ pair in SNLI with an agreed gold label, we replace the text premise with its corresponding Flickr30k image, resulting in a $(P_{image}, H_{text})$ pair in SNLI-VE. 
Figures~\ref{fig:SNLI-VE-eg1} and \ref{fig:SNLI-VE-more-egs} illustrate examples from the SNLI-VE dataset. 
SNLI-VE naturally meets the aforementioned {\em criterion 1} and {\em criterion 2}. Each image in SNLI-VE are real-world ones and is associated with distinct labels given different hypotheses. Furthermore, Flickr30k and SNLI are well-studied datasets, allowing the community to focus on the new task that our paper introduces, rather 
than spending time familiarizing oneself with the idiosyncrasies of a new dataset.

\begin{table}[t]
\centering
\begin{tabular}{lcccc}
\toprule
\toprule
& \textbf{Training} & \textbf{Validation} & \textbf{Testing} \\
\midrule
\textbf{\#Image} & 29,783 & 1,000 & 1,000  \\
\textbf{\#Entailment} & 176,932 & 5,959 & 5,973 \\
\textbf{\#Neutral} & 176,045 & 5,960 & 5,964 \\
\textbf{\#Contradiction} & 176,550 & 5,939 & 5,964 \\
\textbf{Vocabulary Size} & 29,550 & 6,576 & 6,592  \\ 
\bottomrule
\bottomrule
\\
\end{tabular}
\caption{\bf SNLI-VE dataset}
\label{table:dataset}
\vspace{-0.2in}
\end{table}

A sanity check is applied to SNLI-VE dataset partitions in order to guarantee {\em criterion 3}. We notice the original SNLI dataset partitions does not consider the arrangement of the original caption images. 
If SNLI-VE directly adopts the original partitions from SNLI, all images in validation or testing partitions also exist in the training partitions, violating {\em criterion 3}. 
To amend this, we disjointedly partition SNLI-VE by images following the partition in \cite{gong2014improving} and make sure instances with different labels are of similar numbers across training, validation, and testing partitions as shown in Table~\ref{table:dataset}.

Regarding {\em criterion 4}, since SNLI has already been extensively studied, we are aware that there exists a hypothesis-conditioned bias in SNLI as recently reported by Gururangan \etal ~\cite{gururangan2018annotation}.
Though the labels in SNLI-VE are distributed evenly across dataset partitions, SNLI-VE still inevitably suffers from this bias inherently. Therefore, we provide a hypothesis-only baseline in Section~\ref{subsubsec:hypothesis-only} to serve as a performance lower bound.

\subsection{SNLI-VE and VQA Datasets} 

\begin{figure}[h]
  \centering
  \includegraphics[width=\linewidth]{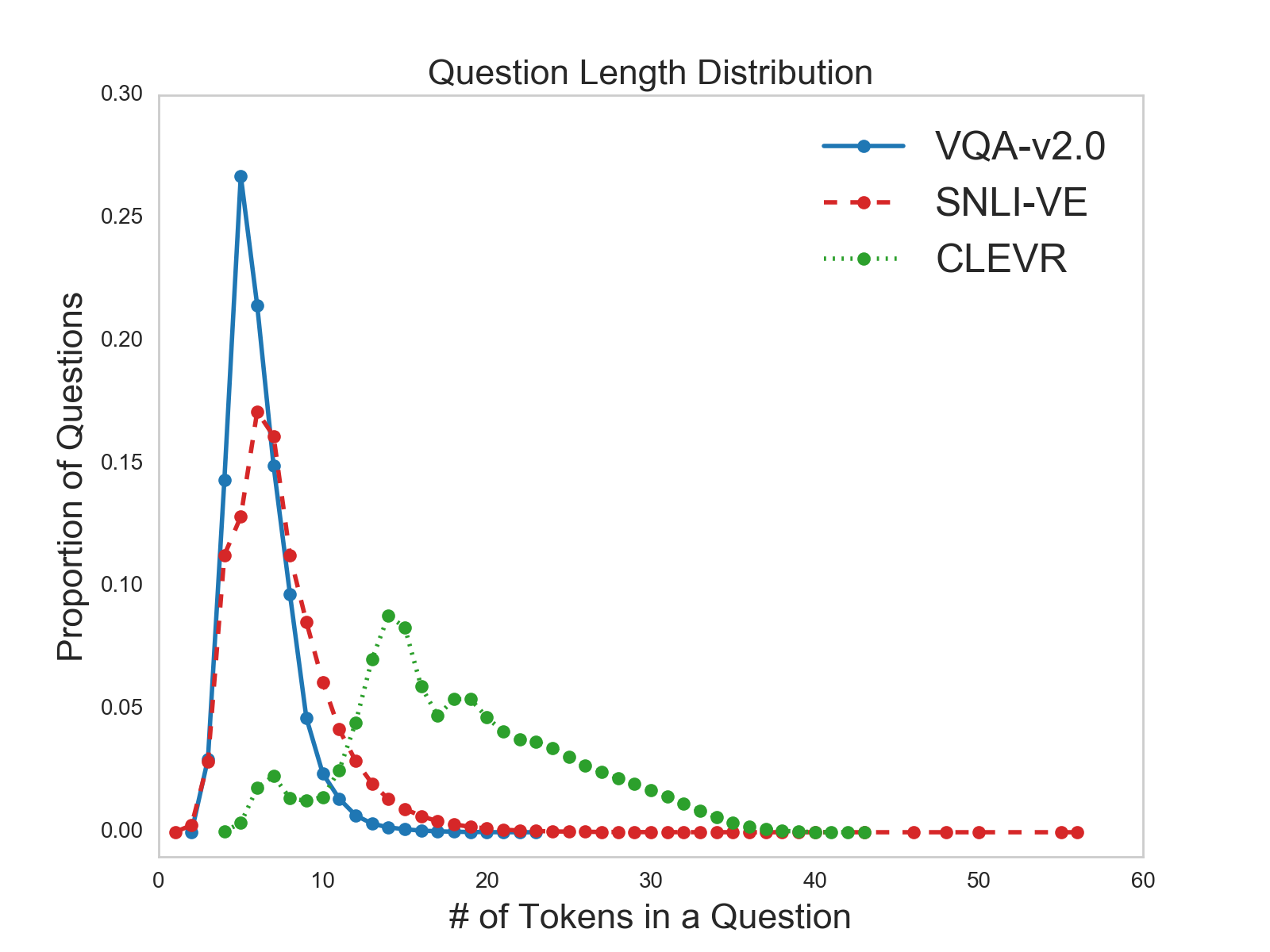}
  \caption{\bf Question Length Distribution}
  \label{fig:SNLI-VE-sentence-length}
\end{figure}


\begin{table}[t]
\centering
\begin{tabular}{lcccc}
\toprule
\toprule
& \textbf{SNLI-VE} & \textbf{VQA-v2.0} & \textbf{CLEVR} \\
\midrule
\textbf{Partition Size:} \\
Training & 529,527 & 443,757 & 699,989  \\
Validation & 17,858 & 214,354 & 149,991  \\
Testing & 17,901 & 555,187 & 149,988  \\
\midrule
\textbf{Question Length:} \\
Mean & 7.4 & 6.1 & 18.4  \\
Median & 7.0 & 6.0 & 17.0  \\
Mode & 6 & 5 & 14  \\
Max & 56 & 23 & 43  \\
\midrule
\textbf{Vocabulary Size} & 32,191 & 19,174 & 87  \\
\bottomrule
\bottomrule
\\
\end{tabular}
\caption{\bf Dataset Comparison Summary}
\label{table:dataset-summary}
\vspace{-0.2in}
\end{table}

We further compare our SNLI-VE dataset with the two widely used VQA datasets, VQA-v2.0 and CLEVR. The comparison focuses on the {\em questions} (for SNLI-VE dataset, we consider a hypothesis as a question). Table~\ref{table:dataset-summary} is a statistical summary about the questions from three datasets. Before generating Table~\ref{table:dataset-summary}, questions are prepossessed by three steps: {\em i)} split into words, {\em ii)} lower case all words, {\em iii)} removing punctuation symbols \{`'``'',.-?!\}. Figure~\ref{fig:SNLI-VE-sentence-length} depicts a detailed question length distribution.

According to Table~\ref{table:dataset-summary}, among the three datasets, our SNLI-VE dataset, which contains the smallest total number of questions (summing up training, validation and testing partitions), has the largest vocabulary size. The maximum question length in SNLI-VE is 56, which is the largest among these three datasets, and represents real-world descriptions. Both the mean and median lengths are larger than VQA-v2.0 dataset. The question length distribution of SNLI-VE, as shown in Figure~\ref{fig:SNLI-VE-sentence-length}, is quite heavy-tailed in contrast to the others.
These observations indicate that the text in SNLI-VE may be difficult to handle compared to VQA-v2.0 for certain models.  As for CLEVR dataset, even though most sentences are much longer than SNLI-VE as shown in Figure~\ref{fig:SNLI-VE-sentence-length}, the vocabulary size is only 87.  We believe this is due to the synthetic nature of CLEVR, which also indicates models that achieve high-accuracy on CLEVR may not be able to generalize to our SNLI-VE dataset.

\section{EVE: Explainable Visual Entailment System}\label{sec:EVE}
\label{sec:eve}

\begin{figure*}[h]
  \centering
  \includegraphics[width=\linewidth]{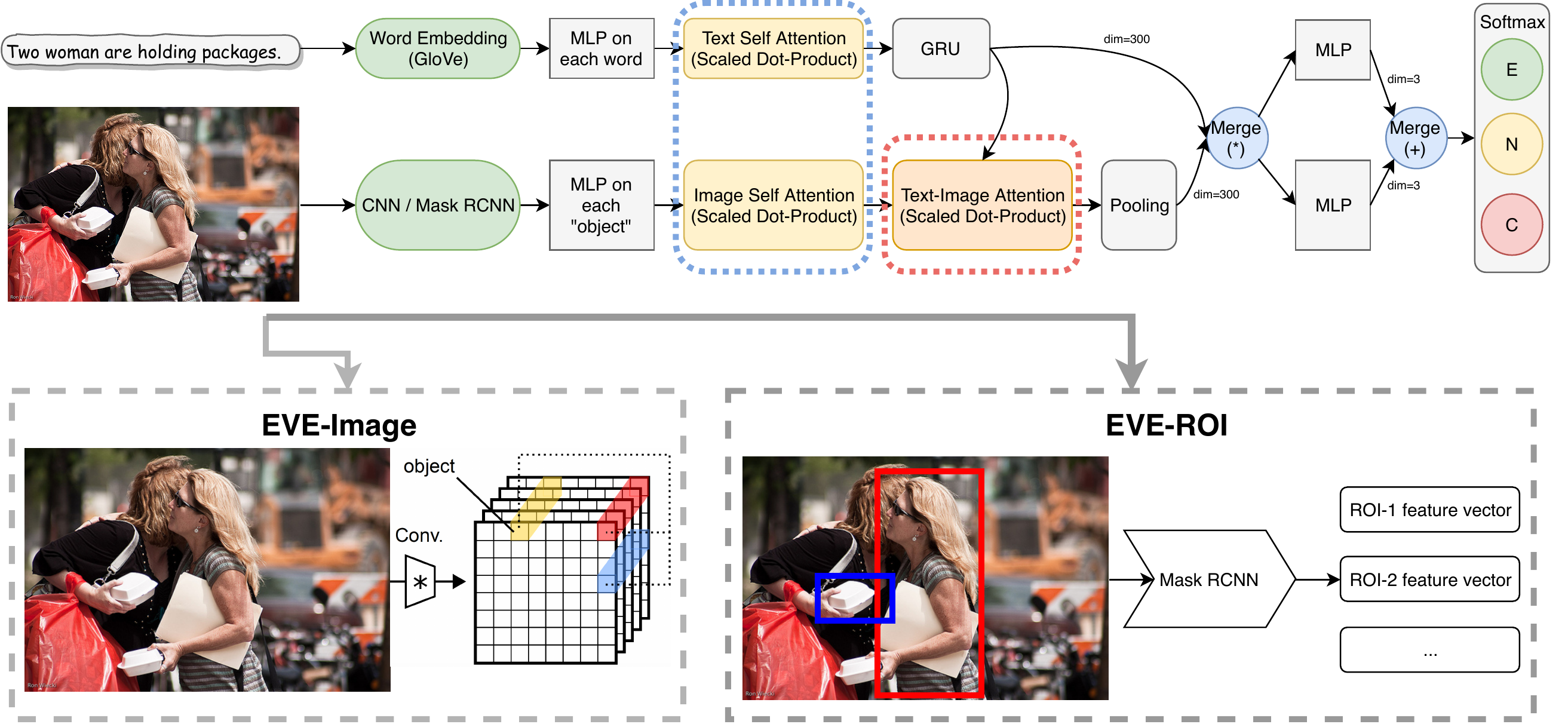}
  \caption{\bf Our model EVE combines image and ROI information to model fine-grained cross-modal information}
  \label{fig:EVE}
  \vspace{-0.2in}
\end{figure*}

The design of our explainable VE architecture, as shown in Figure~\ref{fig:EVE}, is based on the Attention Top-Down/Bottom-Up model discussed later in Subsection~\ref{subsubsec: Attention-Top-Down-Bottom-Up}, which is the winner of VQA Challenge, 2017. 
Similar to the Attention Top-Down/Bottom-Up, our EVE architecture is composed of a text and an image branch.
The text branch extracts features from the input text hypothesis $H_{text}$ through an RNN.
The image branch generates image features from $P_{image}$.
The features produced from the two branches are then fused and projected through fully-connected (FC) layers towards predicting the final conclusion.
The image features can be configured to take the feature maps from a pre-trained convolutional neural network (CNN) or ROI-pooled image regions from a region of interest (ROI) proposal network (RPN).

We build two model variants, \emph{EVE-Image} and \emph{EVE-ROI}, for image and ROI features, respectively.
EVE-Image incorporates a pre-trained ResNet101 ~\cite{he2016deep}, which generates $k$ feature maps of size $d \times d$. 
For each feature map position, the feature vector across all the $k$ feature maps is considered as an \emph{object}.
As a result, there are a total number of $d \times d$ objects of feature size $k$ for an input image. 
In contrast, the EVE-ROI variant takes ROIs as objects extracted from a pre-trained Mask R-CNN ~\cite{MaskRCNNPy}.

In order to accurately solve this cross-model VE task, we need: 
both a mechanism to identify the salient features in images and text inputs and a cross-modal embedding to effectively learn the image-text interactions, which are addressed by employing 
\emph{self-attention} and 
\emph{text-image attention}
techniques in the EVE model respectively. 
We next describe the design and implementation of the mechanisms in EVE model.

\subsection{Self-Attention}\label{sec:eve-self-attention}

EVE utilizes self-attention ~\cite{vaswani2017attention} in both text and image branches as highlighted with dotted blue frame in Figure~\ref{fig:EVE}. 
Since the hypothesis in SNLI-VE can be relatively long and complex, self-attention helps focus on important keywords in a sentence that relate to each other.
The text branch applies self-attention to the projected word embeddings from a multi-layer perceptron (MLP). It is worth noting that although word embeddings, either from GloVe or other existing models, may be fixed, the MLP transformation is able to be trained to generate adaptive projected word embeddings. 
Similarly, the image branch applies the self-attention to projected image regions either from the aforementioned feature maps or ROIs in expectation of capturing the hidden relations between elements in the same feature space.

Specifically, we use the scaled dot product (SDP) attention in ~\cite{vaswani2017attention} to capture this hidden information:
\begin{equation}
Att_\mathrm{sdp} =  \mathrm{softmax}(\frac{RQ^{T}}{\sqrt{d_k}}))
\label{eq:SDP-attention}
\end{equation}
\begin{equation}
Q_{Att} =  Att_\mathrm{sdp}Q
\label{eq:SDP-attention}
\end{equation}
where $Q \in \mathbb{R}^{M \times d_k}$ is the \emph{query} feature matrix and
$R \in \mathbb{R}^{N \times d_k}$ is the \emph{reference} feature matrix. 
$M$ and $N$ represent the number of features vectors in matrix $Q$ and $R$ respectively, 
and $d_k$ denotes the dimension of each feature vector. 
$Att_\mathrm{sdp} \in \mathbb{R}^{N \times M}$ is the resulting attention mask for $Q$ given $R$.
Each element $a_{ij}$ in $Att_\mathrm{sdp}$ represents how much weight (before scaled by $\frac{1}{\sqrt{d_k}}$ and normalized by softmax) the model should put on each query feature vector $q_{j \in \{1, 2, \dots, M \}} \in \mathbb{R}^{d_k}$ in $Q$ w.r.t. each reference feature vector $r_{i \in \{1, 2, \dots, N \}} \in \mathbb{R}^{d_k}$ in $R$.
The attended query feature matrix $Q_{Att} \in \mathbb{R}^{N \times d_k}$ is the weighted and fused version of the original query feature matrix $Q$, calculated by the matrix dot product between the attention mask $Att_\mathrm{sdp}$ and the query feature matrix $Q$.
Note that for the self-attention, the query matrix $Q \in \mathbb{R}^{M \times d_k}$ and the ``reference'' matrix $R \in \mathbb{R}^{N \times d_k}$ are the same matrix. 

\subsection{Text-Image Attention}\label{sec:eve-text-image-attention}

Multi-modal tasks such as phrase grounding~\cite{chen2017query} demonstrate that high-quality cross-modal feature interactions improve the overall performance.
The dotted red frame highlighted area in Figure~\ref{fig:EVE} shows that EVE incorporates the text-image attention to relevant image regions based on the text embedding from the GRU.
The feature interaction between the text and image regions are computed using the same SDP technique introduced in Section~\ref{sec:eve-self-attention}, serving as the attention weights.
The weighted features of image regions are then fused with the text features for further decision making.
Specifically, for the text-image attention, the query matrix $Q \in \mathbb{R}^{M \times d_k}$ is the image features while the ``reference'' matrix $R \in \mathbb{R}^{N \times d_k}$ is the text features. 
Note that although $Q$ and $R$ are from different feature spaces, the dimension of each feature vector is projected to be the same $d_k$ in respective branches for ease of the attention calculation.

\section{Experiments}
\label{sec:eval}

\begin{table*}[h]
\centering
\newcommand{\tabincell}[2]{\begin{tabular}{@{}#1@{}}#2\end{tabular}}
\begin{tabular}{lccccccccc}
\toprule
\toprule

& & \multicolumn{3}{c}{\textbf{Val Acc Per Class (\%)}} & 

\multirow{2}{*}{\tabincell{l}{\textbf{Test Acc} \\ \textbf{Overall (\%)}}} &

\multicolumn{3}{c}{\textbf{Test Acc Per Class (\%)}} \\

\multirow{-2}{*}{\textbf{Model Name}} &

\multirow{-2}{*}{\tabincell{l}{\textbf{Val Acc} \\ \textbf{Overall (\%)}}} & 

\textbf{C}    & \textbf{N} & \textbf{E} & \textbf{} & \textbf{C} & \textbf{N} & \textbf{E} \\

\midrule

\textbf{Hypothesis Only} & 66.68 & 67.54 & 66.90 & 65.60 & 66.71 & 67.60  & 67.71 & 64.83 \\

\textbf{Image Captioning} & 67.83 & 66.61 & 69.23 & 67.65 & 67.67 & 66.25  & 70.69 & 66.08 \\

\textbf{Relational Network} & 67.56 & 67.86 & 67.80 & 67.02 & 67.55 & 67.29 & 68.86 & 66.50 \\

\textbf{Attention Top-Down}  & 70.53  & 70.23 & 68.66 & 72.71 & 70.30 & 69.72  & 69.33 & 71.86 \\

\textbf{Attention Bottom-Up} & 69.34 & \textbf{71.26} & 70.10 & 66.67 & 68.90 & 70.52 & \textbf{70.96} & 65.23  \\

\textbf{EVE-Image*} & \textbf{71.56}  & 71.04  & \textbf{70.55} & 73.10  & \textbf{71.16}  & \textbf{71.56} & 70.52  & 71.39  \\

\textbf{EVE-ROI*}  & 70.81 & 68.55 & 68.78  & \textbf{75.10}  & 70.47 & 67.69 & 69.45 & \textbf{74.25} \\ 

\bottomrule
\bottomrule
\\
\end{tabular}
\caption{\bf Model Performance on SNLI-VE dataset}
\label{table:performance}
\vspace{-0.2in}
\end{table*}

In this section, we evaluate EVE as well as several other baseline models on SNLI-VE.
Most of the baselines are existing or previous SOTA VQA architectures.
The performance results of all models are listed in Table~\ref{table:performance}.

All models are implemented in PyTorch. 
We use the pre-trained GloVe.6B.300D for word embedding~\cite{pennington2014glove}, where 6B is the corpus size and 300D is the embedding dimension.
Input hypotheses are padded to the maximum sentence length in a batch.
Note we do not truncate the sentences because unlike VQA where the beginning of questions typically indicates what is asked about, 
labels of VE task may depend on keywords or small details at the end of sentences.
For example, truncating the hypothesis ``The person who is standing next to the tree and wearing a blue shirt is playing \_\_\_\_\_'' inevitably loses the key detail and changes the conclusion.
In addition, the maximum sentence length in SNLI is 56, which is much larger than 23 in VQA-v2.0 as shown in Table~\ref{table:dataset-summary}.
Always padding to the dataset maximum is not necessarily efficient for training.
As a consequence, we opt for padding to the batch-wise maximum sentence length.

Unless explicitly mentioned, all models are trained using a cross-entropy loss function optimized by the Adam optimizer with a batch size of 64.
We use an adaptive learning rate scheduler which reduces the learning rate whenever no improvement on the validation dataset for a period of time.
The initial learning rate and weight decay are both set to be $1e-4$.
The maximum number of training epochs is set to 100.
We save a checkpoint whenever the model achieves a higher overall validation accuracy.
The final model checkpoint selected for testing is the one with the highest lowest per class accuracy in case the model performance is biased towards particular classes.
The batch size is set as 32 for validation and testing.
In the following, we discuss the details for each baseline.

\subsection{Hypothesis Only} 
\label{subsubsec:hypothesis-only}

This baseline verifies the existing data bias in the SNLI dataset, as mentioned by Gururangan~\etal~\cite{gururangan2018annotation} and Vu~\etal~\cite{vu2018grounded}, by using hypotheses only without the image premise information.

The model consists of a \emph{text processing component} followed by two FC layers. 
The text processing component is used to extract the text feature from the given hypothesis. 
It first generates a sequence of word-embeddings for the given text hypothesis. 
The embedding sequence is then fed into a GRU ~\cite{chung2014empirical} to output the text features of dimension 300.
The input and output dimensions of the two FC layers are [300, 300] and [300, 3] respectively. 

Without any premise information, this baseline is supposed to make a random guess out of the three classes but the resulting accuracy is up to 67\%, implying the existence of a dataset bias.
We do not intend to rewrite the hypotheses in SNLI to reduce the bias but instead, aim at using the premise (image) features to outperform the hypothesis only baseline.

\subsection{Image Captioning} 

Since the original SNLI premises are image captions, a straightforward idea to address VE is to first apply an image caption generator to convert image premises to text premises and then followed by a TE classifier.
Particularly, we adopt the PyTorch tutorial implementation \cite{CaptioningPy} as a caption generator.
A pre-trained ResNet152 serves as the image encoder while the caption decoder is a long short-term memory (LSTM) network.
Once the image caption is generated, the image premise is replaced with the caption and the original VE task is reduced to a TE task.
Similar to the Hypothesis-Only baseline, the TE classifier is composed of two text processing components to extract text features from both the premise and hypothesis.
The text features are fused and go through two FC layers with input and output dimensions of [600, 300] and [300, 3] for the final prediction.

The resulting performance achieves a slightly higher accuracy of 67.83\% and 67.67\% on the validation and testing partitions over the Hypothesis-Only baseline, implying that the generated image caption premise does not improve much.
We suspect that the generated captions may not cover the necessary information in the image as required by the hypothesis to make the correct conclusion.
This is possible in a complex scene where exhaustive enumeration of captions may be needed to cover every detail potentially described by the hypothesis.

\subsection{Relational Network} 

The Relational Network (RN) baseline is based on ~\cite{santoro2017simple} which is proposed to tackle the CLEVR dataset with high accuracy.
There are an image branch and a text branch in the model. 
The image branch extracts image features in a similar manner as EVE, as described in Section~\ref{sec:EVE}, but without self-attention.
The text branch generates the hypothesis embedding through an RNN.
The highlight of RN is to capture pairwise feature interactions between image regions and the text embedding.
Each pair of image region feature and question embedding goes through an MLP.
The final classification takes the element-wise sum over the MLP output for each pair as input.

Despite the high accuracy on the synthetic dataset CLEVR, RN only achieves a marginal improvement on SNLI-VE at the accuracy of 67.56\% and 67.55\% on the validation and testing partitions.
This may be attributed to the limited representational power of RN that fails to produce 
effective cross-modal feature fusion 
of the natural image premises and the free-form text hypothesis input from SNLI-VE.

\subsection{Attention Top-Down and Bottom-Up} \label{subsubsec: Attention-Top-Down-Bottom-Up}

We consider the Attention Top-Down and Attention Bottom-Up baselines based on the winner of VQA challenge 2017~\cite{anderson2018bottom}.  
Similar to the RN baseline, there is an image branch and a text branch.
The difference between the image branches in Attention Top-Down and Attention Bottom-Up is similar to our EVE.
The image features of Attention Top-Down come from the feature maps generated from a pre-trained CNN.
As for Attention Bottom-Up, the image features are the top 10 ROIs extracted from a pre-trained Mask-RCNN implementation ~\cite{he2017mask}.
No self-attention is applied in both image and text branches.
Moreover, the text-image attention is implemented by feeding the concatenation of both image and text features into an FC layer to derive the attention weights rather than using SDP as described in Section~\ref{sec:eve-self-attention}.
Then the attended image features and text features are projected separately and fused by dot product.
The fused features go through two different MLPs.
The element-wise sum of both MLP output serves as the final features for classification.

The SOTA VQA winner model, Attention Top-Down, achieves an accuracy of 70.53\% and 70.30\% on the validation and testing partitions respectively, implying cross-modal attention is the key to effectively leveraging image premise features.
The Attention Bottom-Up model using ROIs also achieves a good accuracy of 69.34\% and 68.90\% on the validation and testing partitions.
The reason why Attention Bottom-Up performs worse than Attention Top-Down could be possibly due to lack of background information in ROI features and ROI feature quality. 
It is not guaranteed that those top ROIs cover necessary details described by the hypothesis.
However, even with more than 10 ROIs, we observe no significant improvement in performance.

\subsection{EVE-Image and EVE-ROI} 

The details of our EVE architecture have been described in Section~\ref{sec:EVE}.
EVE-Image achieves the best performance of 71.56\% and 71.16\% accuracy on the validation and testing partitions respectively.
The performance of EVE-ROI is similar, with an accuracy of 70.81\% and 70.47\%, possibly suffering from similar issues as the Attention Bottom-Up model.
However, the improvement is likely due to the introduction of self-attention and text-image attention through SDP that potentially captures the hidden relations in the same feature space and better attended cross-modal feature interaction.

\begin{figure}[h]
 \centering
 \begin{minipage}[t]{.45\columnwidth}
  
 \includegraphics[width=\linewidth]{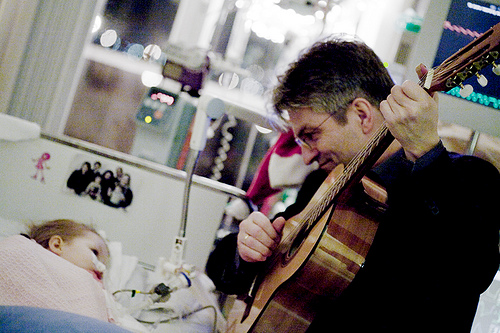}
 \end{minipage}\hfill
 \begin{minipage}[t]{.45\columnwidth}
  \centering
 \includegraphics[width=\linewidth]{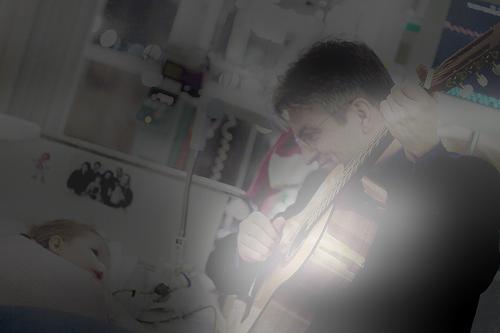}
 \end{minipage}\hfill
 \caption{An attention visualization for EVE-Image}
 \label{fig:EVE-image-attention}
 \end{figure}
 
\begin{figure}[h]
 \centering
 \begin{minipage}[t]{.45\columnwidth}
  \centering
 \includegraphics[width=\linewidth]{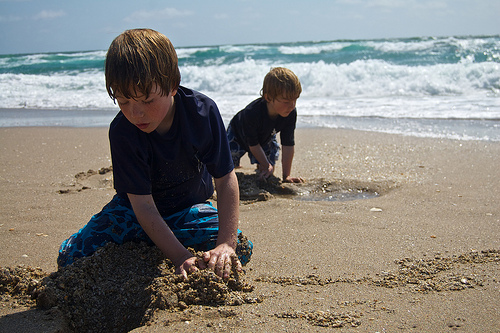}
 \end{minipage}\hfill
 \begin{minipage}[t]{.45\columnwidth}
  \centering
 \includegraphics[width=\linewidth]{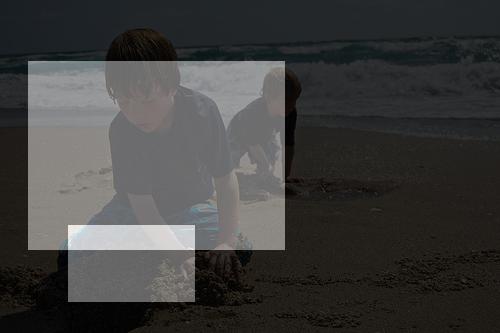}
 \end{minipage}\hfill
 \caption{\bf An attention visualization for EVE-ROI}
 \label{fig:EVE-ROI-attention}
 \vspace{-0.2in}
 \end{figure}
 
\noindent {\bf Attention Visualization.} 
The explainability of EVE is attained using attention visualizations in the areas of interest in the image premise given the hypothesis. 
Figure~\ref{fig:EVE-image-attention} and \ref{fig:EVE-ROI-attention} illustrate two visualization examples of the \emph{text-image attention} from EVE-Image and EVE-ROI respectively. 
The image premise of the EVE-Image example is shown on the left of Figure~\ref{fig:EVE-image-attention}, and the corresponding hypothesis is \emph{``A human playing guitar''}.
On the right of Figure~\ref{fig:EVE-image-attention}, our EVE-Image model successfully attends to the guitar area, leading to the correct conclusion: \emph{entailment}. 
In Figure~\ref{fig:EVE-ROI-attention}, our EVE-ROI focuses on the children and the sand area in the image premise, leading to the \emph{contradiction} conclusion for the given hypothesis \emph{``Two children are swimming in the ocean.''}

\subsection{Discussion}
\label{subsec:discussions}

In this section, we discuss why existing VQA and CLEVER models have modest performs over SNLI-VE dataset
and the possible future directions based on our experience. 
VQA models are not trained to distinguish fine-grained information. Furthermore,
with the same image present across all the three classes in the SNLI-VE dataset,
SNLI-VE removes any bias that may originate from just the image premise information
and an effectively fused representation is important for high accuracy.
Furthermore, models that provide good performance on CLEVR may not work on SNLI-VE
since these models have rather simplistic image processing pipelines, 
often with a couple of convolutional layers that may be sufficient to process synthetic images but works poorly on real images. More importantly, the sentences are not synthetic in the SNLI-VE dataset. 
As a result, building compositional reasoning modules over SNLI-VE hypotheses is out of reach for existing models.

To effectively address SNLI-VE, we believe three approaches can be beneficial. 
First, using external knowledge beyond pre-trained models and/or visual entity extraction can be beneficial. 
If the external knowledge can provide information allowing the model to learn relationships between the entities that
may be obvious to humans but difficult or impossible to learn from the dataset (such as ``two women in the image are sisters''),
it will improve the model performance over SNLI-VE.

Second, it is possible for the hypothesis to contain multiple class labels assigned to its different entities or relationships w.r.t. the premise. 
However, SNLI-VE lacks annotations for localizing the labels to specific entities in the hypothesis (e.g. as is often provided in synthetic datasets like bABi~\cite{weston2015towards}). 
Since the hypothesis can be broken down into individual entities and relationships between pairs of entities, providing fine-grained labels for each target in the hypothesis likely facilitates strongly-supervised training.

Finally, a third possible approach is to build effective attention based models as
done in TE that encodes the sentences together 
and align similar words in hypothesis and premise instead
of a late-fusion of separately encoded modalities. Hence,
the active research on visual grounding can benefit addressing
the SNLI-VE task.

\section{Conclusion}
\label{sec:conclusion}

We introduce a novel task, visual entailment, that requires fine-grained reasoning over the image and text.
We build the SNLI-VE dataset for VE using real-world images from Flickr30k as premises, and the corresponding text hypotheses from SNLI.
We then develop the EVE architecture to address VE and evaluate against multiple baselines, including existing SOTA VQA based models.
We expect more effort to be devoted to generating fine-grained VE annotations for large image datasets such as the Visual Genome~\cite{krishnavisualgenome} and Open Images Dataset~\cite{openimages} as well as improved models on fine-grained visual reasoning.

\section*{Acknowledgments}
Ning Xie and Derek Doran were supported by the Ohio Federal Research Network project \textit{Human-Centered Big Data}. 
Any opinions, findings, and conclusions or recommendations expressed in this article are those of the author(s) and do not necessarily reflect the views of the Ohio Federal Research Network.

{\small
\bibliographystyle{ieee}
\bibliography{egbib}
}

\clearpage
\onecolumn
\section*{Supplementary: Additional Examples}

\begin{figure*}[h]
\centering
  \includegraphics[width=\linewidth]{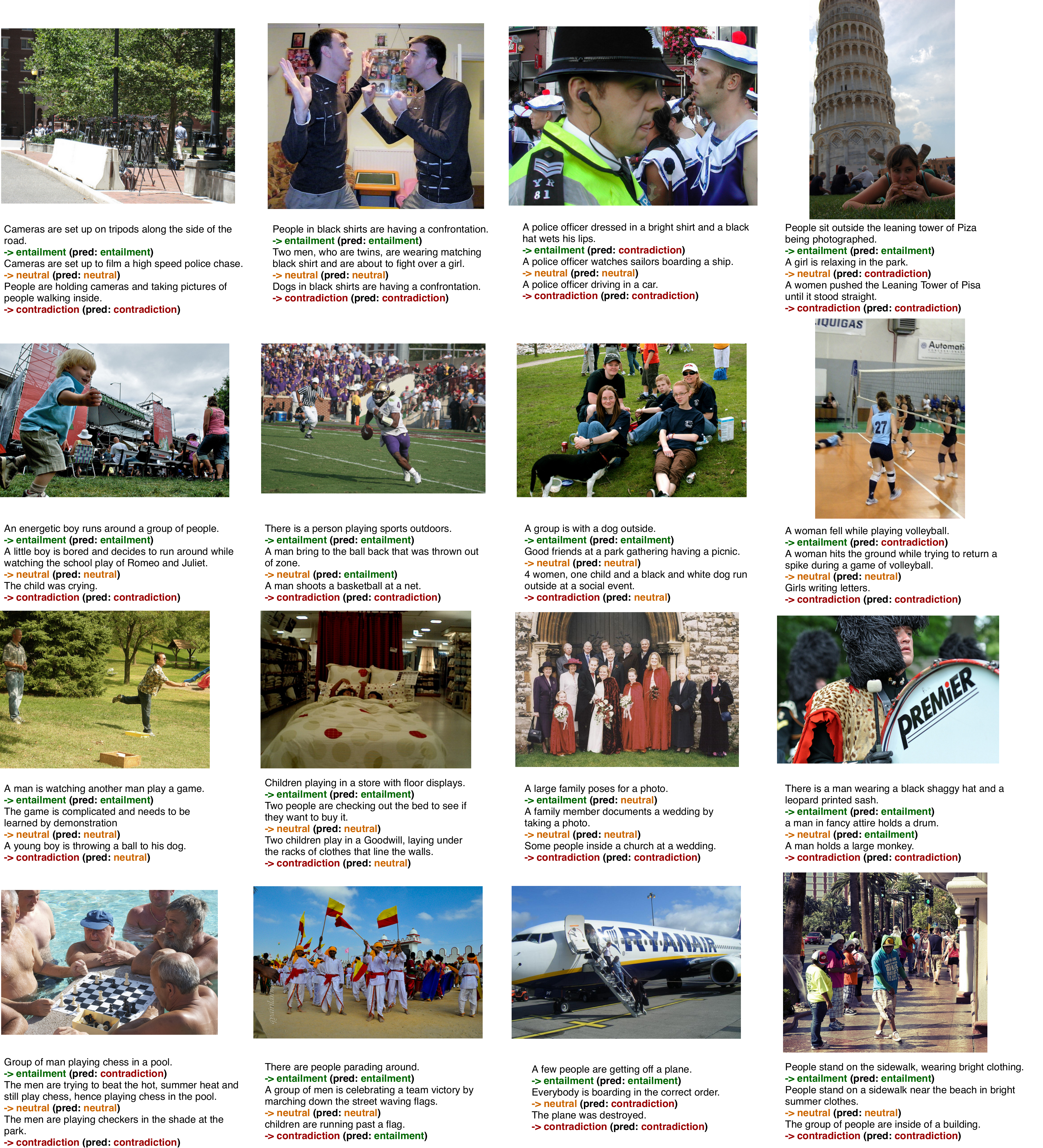}
\caption{Random examples from SNLI-VE with prediction results from our best-performed EVE-Image}
\label{fig:SNLI-VE-random-egs}
\end{figure*}

Figure~\ref{fig:SNLI-VE-random-egs} shows random examples from SNLI-VE with predictions from our EVE-Image.
Each example consists of an image premise and three selected hypotheses of different labels. 
Note that for each image premise, the total number of hypotheses are not limited to three.

\end{document}